\documentclass[10pt,twocolumn,letterpaper]{article}
\usepackage{cvpr}

\definecolor{cvprblue}{rgb}{0.21,0.49,0.74}
\usepackage[pagebackref,breaklinks,colorlinks,allcolors=cvprblue]{hyperref}
\usepackage[accsupp]{axessibility} 
\def\confYear{2026}

\title{Ninja Codes: Neurally Generated Fiducial Markers for Stealthy 6-DoF Tracking}

\author{
\setlength{\tabcolsep}{20pt}
\begin{tabular}[t]{ccc}
Yuichiro Takeuchi & Yusuke Imoto & Shunya Kato\\
Sento & Osaka University & Kyoto University\\
\end{tabular}\\
[3.2ex]
\href{https://sento.net/research/ninjacodes}{\normalsize\tt sento.net/research/ninjacodes}
\vspace{-2pt}
}

\begin{document}

\twocolumn[{%
\renewcommand\twocolumn[1][]{#1}%
\maketitle
\begin{center}
    \centering
    \captionsetup{type=figure}
    \includegraphics[width=1.0\linewidth]{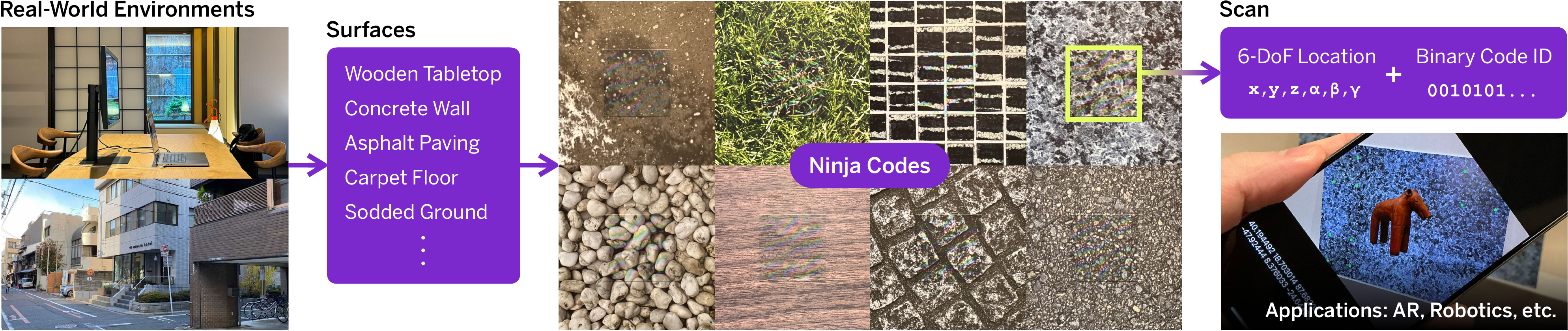}
    \captionof{figure}{We present Ninja Codes, inconspicuous fiducial markers that can blend into various real-world environments. An encoder neural network converts photos of surface textures into Ninja Codes, which can be printed out using off-the-shelf color printers. The codes can be used to offer real-time 6-DoF tracking, in the same manner as conventional fiducial markers.}
\end{center}%
}]

\begin{abstract}
In this paper we describe Ninja Codes, neurally generated fiducial markers that can be made to naturally blend into various real-world environments. An encoder network converts arbitrary images into Ninja Codes by applying visually modest alterations; the resulting codes, printed and pasted onto surfaces, can provide stealthy 6-DoF location tracking for a wide range of applications including robotics and augmented reality. Ninja Codes can be printed using standard color printers on regular printing paper, and can be detected using any device equipped with a modern RGB camera and capable of running inference. Through experiments, we demonstrate Ninja Codes' ability to provide reliable location tracking under common indoor lighting conditions, while successfully concealing themselves within diverse environmental textures. We expect Ninja Codes to offer particular value in scenarios where the conspicuous appearance of conventional fiducial markers makes them undesirable for aesthetic and other reasons.
\end{abstract}

\section{Introduction}
Precise, reliable location tracking is a foundation of numerous computing applications such as robotics and augmented reality. Though a range of solutions exist for this task, old-fashioned fiducial markers --- 2D graphical images that can be printed on paper and detected using computer vision --- still retain broad appeal due to the combination of low cost, ease of deployment, and robust performance under a broad array of conditions. However, the typically eye-catching appearance of fiducial markers makes their use unsuitable in some situations, especially those where aesthetic concerns carry weight. Hence, despite their widespread use in select environments such as research labs and warehouses, adoption in non-professional settings such as residential homes remains limited. Inconspicuous fiducial markers that seamlessly blend into environments can potentially introduce inexpensive, accurate location tracking to novel contexts.

In this paper we describe Ninja Codes (Figure 1), a new class of fiducial markers that are able to conceal themselves within various real-world environmental textures. Building upon prior work in the field of deep steganography \cite{baluja2017}, we train a neural network that turns arbitrary images into functional fiducial markers through subtle visual alterations. By creating Ninja Codes from photographs of wooden tabletops, tiled walls, asphalt paving, etc., we can obtain discreet fiducial markers that can be installed throughout environments in visually unobtrusive ways. The codes can be detected by a wide range of digital devices equipped with an RGB camera and capable of running inference --- a category of devices which includes most modern smartphones --- and can be printed out using commonly available equipment.

The Ninja Codes creation/detection pipeline comprises multiple neural network modules, which are trained jointly using an end-to-end process. Differentiable noise functions are applied at two different stages of the training process, to increase resistance to real-world perturbations such as camera noise and specular reflection. We evaluated the performance of Ninja Codes through a series of experiments, and the results demonstrate their ability to offer reliable 6-DoF tracking under normal indoor lighting conditions, while effectively concealing themselves from human detection.

The key contributions of this paper are as follows:
\begin{itemize}
\item{Introduces Ninja Codes, a new category of fiducial markers that builds upon deep steganography techniques}
\item{Reports experimental results validating the codes' capacity as a reliable and discreet location tracking solution}
\item{Identifies outstanding issues and limitations regarding the technology, and maps out avenues for improvement}
\end{itemize}

A brief note on our terminology --- in this paper, we use the term {\it location tracking} as a synonym for 6-DoF tracking of both position ($x, y, z$) and pose ($\alpha, \beta, \gamma$). Also, when we use the term {\it detection} with no qualifiers, we refer to the entire process of finding Ninja Codes within scenes, obtaining their corner coordinates, and retrieving their respective IDs.

\section{Related Work}

\subsection{Fiducial Markers}

Modern variations of two-dimensional fiducial markers first appeared in the early 1990s \cite{gatrell1992}. Since then, they have seen sustained use as a reliable and inexpensive location tracking technology, particularly in industrial and/or professional settings such as factories, warehouses, film production studios, research labs, etc. Examples of fiducial markers currently in wide use include ARTag \cite{fiala2005}, AprilTag \cite{olson2011}, and ArUco \cite{ramirez2018}, all of which encode relatively modest amounts of data (e.g., 36 bits) in simple square images that can be printed using standard printers. Apart from a small number of examples that employ non-rectilinear shapes \cite{bergamasco2011, jorda2007} or use non-grayscale color \cite{degol2017}, most markers are monochrome and have grid-like appearances reminiscent of QR codes but simpler due to their smaller data capacities. (Fiducial markers only need to encode short IDs to differentiate them from other markers. For example, a marker with 36-bit data capacity can represent $2^{36}$ $\approx$ 68.7 billion unique IDs.)

Most fiducial markers are detected using handcrafted algorithms. However, a string of work within the past decade has explored the use of deep learning, either to detect existing markers like ArUco \cite{hu2019} or to develop wholly new fiducial markers that are both created and detected using neural networks \cite{grinchuk2016, peace2020}. A notable advantage of such neurally generated markers is that through strategic design of model architecture, training pipeline, etc., it is possible to develop markers with unique attributes that would be difficult to design via conventional means. DeepFormableTag \cite{yaldiz2021} is one example of such work, which can retain effectiveness even when attached to non-planar surfaces.

Development of inconspicuous fiducial markers that naturally blend into environments is a topic that has received relatively little attention, possibly due to the fact that fiducial markers are commonly used in settings where aesthetic concerns are often deemed secondary. Proposed techniques include the use of infrared-absorbing ink \cite{willis2013} or fluorescent polymers \cite{dogan2023} to create markers that are visible to specialized cameras but not to humans, and embedding hidden air pockets inside 3D printed objects that can be read out as markers by camera/projector units \cite{li2017}. Outside of fiducial markers for location tracking, efforts have been made \cite{yang2016, xu2019, xu2021} to design aesthetically pleasing markers for data transmission, i.e., more attractive alternatives to QR codes.

Regular images can be appropriated as fiducial markers, in cases where only a small, predefined set of images needs to serve as markers. Augmented reality apps that use corporate logos or product packages to trigger effects are a prime example of this. (A classic, but still widely used method for implementing such systems is to track feature vectors \cite{lowe2004}.) At small scales, we can adapt such techniques to realize discreet fiducial markers. However, they cannot accommodate scenarios where a large number of mutually distinguishable markers need to be placed throughout an environment --- a necessary setup to offer robust location tracking in any reasonably sized space.

\begin{figure*}[t]
  \centering
  \includegraphics[width=1.0\linewidth]{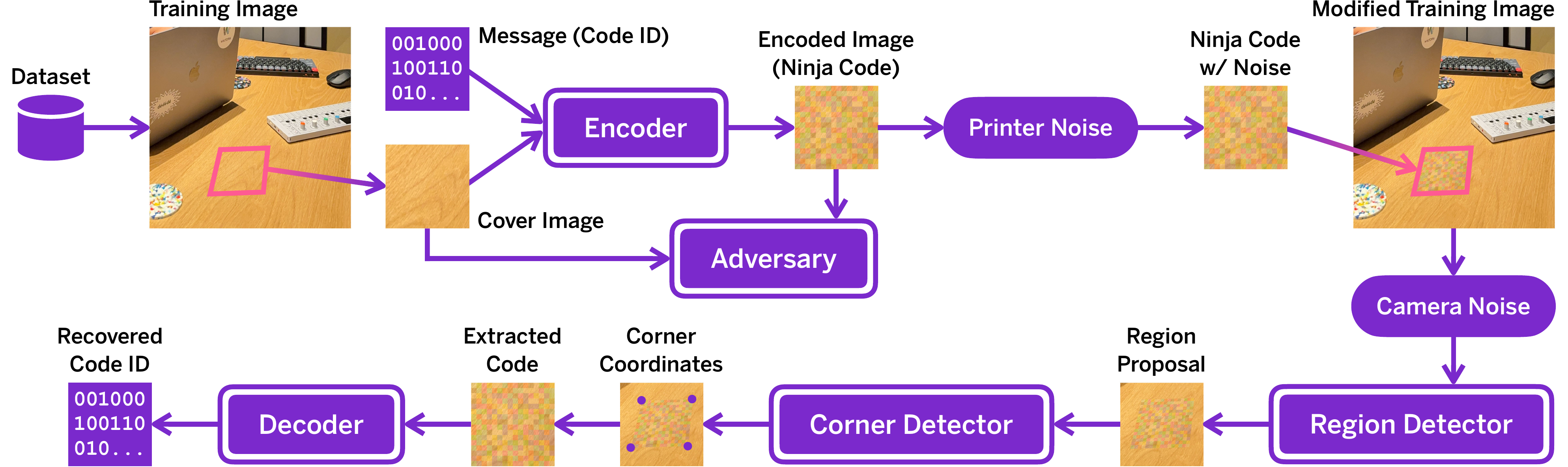}
  \caption{Ninja Codes end-to-end training pipeline. A total of five modules are trained simultaneously: encoder, decoder, region detector, corner detector, and adversary. A series of differentiable noise functions are applied during the training process, to simulate perturbations.}
\end{figure*}

\subsection{Other Location Tracking Technologies}

There exists a number of location tracking technologies besides fiducial markers. Many techniques involve the use of external hardware such as WiFi basestations \cite{bahl2000}, ultrasound beacons \cite{priyantha2000}, or infrared camera/illuminator combinations \cite{pintaric2007}, and offer reliable, accurate tracking albeit with high costs of setup and operation which limit their utility outside of select professional use cases. There also exist techniques that require no external hardware (or even markers), for example inertial sensing \cite{brossard2020}, SLAM \cite{cadena2016}, and so-called visual positioning systems (VPS) developed by corporations such as Google. Although the advantages of such self-contained solutions are undeniable, fiducial markers still retain a significant edge in many real-world scenarios; SLAM and VPS struggle in environments with repetitive or feature-poor surfaces, and inertial sensing using consumer-grade hardware typically only yields middling precision.

A wide variety of applications rely on the availability of precise location tracking. Augmented reality is a representative example, where the accuracy of location tracking has direct correlation to the breadth and quality of visual effects that can be implemented. Advanced effects such as diminished reality \cite{mann2002, herling2010} and freeform environmental transformations \cite{takeuchi2012} are only possible when precise 6-DoF tracking is available. Robotics is another example; research demos such as drone-based construction \cite{gramazio2012} showcase what could be accomplished with precision tracking in this area.

\subsection{Deep Steganography}

Steganography refers to the practice of surreptitiously embedding arbitrary messages within various media, most often images and video; deep steganography \cite{baluja2017} is its modern evolution that employs deep learning techniques. HiDDeN \cite{zhu2018} and StegaStamp \cite{tancik2020} are two notable examples of work in this area, both operating on the principle of jointly training encoder/decoder networks using an end-to-end process. Images encoded using these techniques show minimal signs of tampering, and are nearly indistinguishable to the human eye from unprocessed images. The field is attracting growing interest from the research community, leading to variations such as \cite{jia2022} which enables data to be encoded in sub-regions within larger images, opening up new use cases.

Our work borrows heavily from such precedents in deep steganography. While techniques like HiDDeN exclusively concern themselves with the dual processes of encoding and decoding images, we attempt --- taking cues from neural object detection \cite{liu2016, redmon2016, ren2016} --- to also allow encoded images, installed at unknown locations within environments in unknown numbers, to be accurately found and located down to their precise corner coordinates before decoding. To our knowledge, our work presents the first example of neurally generated, inconspicuous fiducial markers built upon deep steganography techniques.

\begin{figure*}[t]
  \centering
  \includegraphics[width=1.0\linewidth]{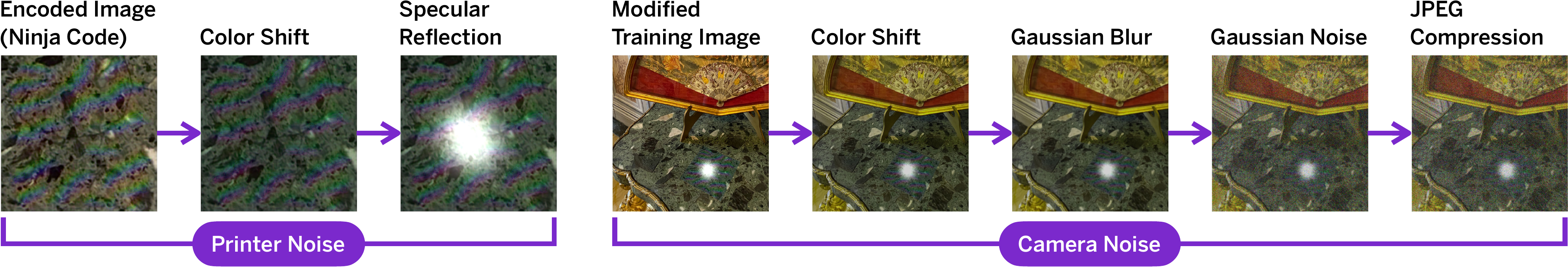}
  \caption{Differentiable noise functions. Perturbations owing to the printing process are simulated using color shift and synthetic specular reflection. Perturbations owing to camera capture are simulated using color shift, Gaussian blur, Gaussian noise, and JPEG compression.}
\end{figure*}

\section{Ninja Codes}

\subsection{Overview}

Figure 2 illustrates the end-to-end pipeline used to train the multiple network modules involved in the creation and detection of Ninja Codes. For each image sampled from the training dataset (with a uniform resolution of 1024$\times$1024), up to four square patches of random sizes are cut out at random locations. The patches are then warped (each vertex is shifted in both $x$ and $y$ directions, within a maximum amplitude set to 15\% of the patch's edge length) and  rotated to form convex quadrilaterals. The quadrilaterals are extracted from the training image, then geometrically rectified and resized into 256$\times$256 square images; borrowing terminology from steganography research, we call these {\it cover images}.

Next, each cover image, along with a randomly assigned 36-bit ID, is given to the encoder which outputs an encoded image (Ninja Code). After applying a series of noise functions intended to simulate perturbations owing to the printing process (e.g., color shift), the Ninja Codes are re-warped and pasted back onto the training image that they had originally been cut out from; the resulting image mimics a real-world scene in which Ninja Codes have been attached onto environmental surfaces. A second round of noise functions is applied to this image, this time intended to simulate perturbations arising from the camera capture process (e.g., optical blur). The image is then resized to 300$\times$300, and given as input to the region detector which outputs square regions each likely to contain a Ninja Code. The results are refined by the corner detector to obtain corner coordinates.

Finally, the located Ninja Codes are geometrically rectified, scaled to 256$\times$256, and given to the decoder to retrieve their IDs. In real-world detection tasks, the position/pose of the capturing device can be computed using a least squares approach, provided that both the 3D world coordinates and physical dimensions of the detected codes are known.

\subsection{Models}

Below are details of the five network modules trained jointly through the end-to-end pipeline.

The {\bf Encoder} takes an RGB cover image and a 36-bit ID as input. The ID is preprocessed using a linear transformation block to produce a tensor with the same dimensions as the cover image (256$\times$256, 3 channels), which is then concatenated with the cover image to create a 6-channel tensor. The resulting tensor is given to a U-Net \cite{ronneberger2015}, whose output is an encoded Ninja Code.

The {\bf Region Detector} is an object detector based on the SSD \cite{liu2016} architecture, which takes a 300$\times$300 image as input and outputs a series of square regions, each deemed to contain a Ninja Code with high probability. As we are only interested in finding Ninja Codes, we use a single-class design with exclusively square-shaped anchors. For the backbone we use VGG \cite{simonyan2015} instead of the more common ResNet \cite{he2016}, prioritizing faster convergence.

The {\bf Corner Detector} is a U-Net followed by a series of fully-connected layers. It takes the SSD region proposal as input, and directly computes the corner points (four sets of $x$, $y$ coordinates) of the enclosed Ninja Code. Alternatively, the final layers can be modified to yield a heatmap. (While heatmaps are the more popular choice \cite{bulat2016} in keypoint detection, we found direct regression to perform better in cases where Ninja Codes partly lie outside the image frame.)

The {\bf Decoder} is a sequence of convolutions followed by a pair of fully-connected layers. Unlike some prior work on deep steganography \cite{tancik2020}, we do not include a spatial transformer \cite{jaderberg2015} to absorb geometric perturbations; the rationale here is that, if the region/corner detectors are performing to expected levels, we can assume the decoder input to already be geometrically rectified.

The {\bf Adversary} is a series of convolutional blocks, followed by averaging pooling and a fully-connected layer. It takes a 256$\times$256 RGB image as input, and outputs its belief (as logit scores) on whether the input is a Ninja Code or an unprocessed cover image.

\subsection{Noise}

Figure 3 illustrates the differentiable noise functions that are applied at two separate stages of the training process.

PRINTER NOISE. The first set of noise functions simulates perturbations owing to the printing method and/or material, and is applied to Ninja Codes immediately after they have been generated by the encoder. We first apply a randomized color shift, where hue ($\pm$0.1), brightness ($\pm$10\%), contrast and saturation ($\pm$15\% each) are all adjusted within predefined ranges. We then overlay a randomly synthesized specular reflection on each code;  this function was informed by the observation that many brands of regular printing paper show moderate degrees of glossiness.

CAMERA NOISE. The second set of noise functions is applied to images given as input to the region detector, and is intended to simulate perturbations arising from the camera capture process. We again perform a randomized color shift (details are identical to the earlier operation), followed by Gaussian blur (with kernel size chosen from 1, 3, 5, and 7px), Gaussian noise (with $\sigma$ sampled from uniform distribution $[0, 0.2]$), and random-quality JPEG compression.

Note that these noise functions have been designed with the assumption that Ninja Codes will be printed on regular paper using standard color printers, and will be captured by common RGB cameras under unexceptional lighting conditions. Accommodating scenarios outside of this assumption will require modified sets of noise functions. For example, while Ninja Codes can theoretically be embedded into surface textures of full-color 3D printed objects, the functions will likely need to be updated to reflect the different, typically limited color space afforded by 3D printers.

\subsection{Loss Function}

We train the modules by minimizing a weighted sum of the following individual losses.

The {\bf Image Loss} ($L_i$) quantifies the visual disparity between Ninja Codes and the original cover images. The loss can be further decomposed as $L_i = L_{px} + L_{ch} + 0.01 * L_{lp}$. Here, $L_{px}$ is the pixel MSE between Ninja Codes and cover images in YUV color space, with each channel weighted to account for sensitivity characteristics of human vision. $L_{ch}$ is a smooth L1 loss intended to ensure that chroma deviations from cover images remain balanced. $L_{lp}$ is the LPIPS learned perceptual similarity metric \cite{zhang2018}. 

The {\bf Regression Loss} ($L_r$) and {\bf Classification Loss} ($L_c$) are the magnitudes of positional error and  classification error, both pertaining to region detector output. As our region detector is SSD-based, we follow \cite{liu2016} and take the smooth L1 loss for the former and cross entropy loss for the latter.

The {\bf Keypoint Loss} ($L_k$) is the degree of error in corner detector output. We take the MSE between predicted corner coordinates and the corresponding ground truth values.

The {\bf Message Loss} ($L_m$) is the degree of error in decoder output. We take the MSE between retrieved and originally assigned 36-bit IDs of Ninja Codes.

The {\bf Adversary Loss} ($L_a$) measures the degree to which the adversary could correctly distinguish between encoded and unencoded images, obtained by giving Ninja Codes as input to the adversary and computing the cross entropy loss between its output logits and a target tensor.

To summarize, the total loss is expressed as follows, with $w_i$, $w_r$, $w_c$, $w_k$, $w_m$, $w_a$ denoting respective weights.
\begin{align*}
L & = w_i L_i + w_r L_r + w_c L_c + w_k L_k + w_m L_m + w_a L_a
\end{align*}

Note that while $L$ includes the adversary loss, this loss is not used to train the adversary itself --- it is instead trained separately, using the reverse of $L_a$ as the loss function.

\subsection{Training}

We performed training using two NVIDIA 3090 GPUs. For all training we used the Adam \cite{kingma2015} optimizer, with learning rate set to $1.0e^{-5}$, weight decay set to zero, and ($\beta_1$, $\beta_2$, $\epsilon$) set to ($0.9$, $0.999$, $1.0e^{-8}$), respectively.

DATASET. We prepared a dataset containing 48,000 images in total, consisting of 43,300 images from the general-purpose COCO (Common Objects in Context) dataset \cite{lin2014}, and 4,700 images from the DTD texture image dataset \cite{cimpoi2014}. Each image was resized and cropped into a uniform size of 1024$\times$1024. We also prepared a smaller dataset containing 9,600 images for validation, similarly consisting of images taken from COCO and DTD.

TRAINING (PHASE 1). We found that splitting training into two separate phases leads to more reliable convergence and predictable results. In the first phase, we ignore image loss, message loss, and adversary loss and focus solely on improving region and corner detection performance, setting weights to $w_i$=$w_m$=$w_a$=$0$, $w_r$=$w_c$=$1.0$, and $w_k$=$5.0e^3$. We train for a total of 20 epochs, after which the encoder will have learned to generate vivid, striped Ninja Codes (Figure 4, top left). The pattern of the stripes is an emergent feature, and manifests differently each time training is performed.

TRAINING (PHASE 2). In the second phase we attempt to minimize all losses, training for 60 epochs with weights set to $w_i$=$w_r$=$w_c$=$1.0$, $w_k$=$5.0e^3$, $w_m$=$10.0$, and $w_a$=$0.001$ initially, and progressively increasing $w_i$ to $100.0$ over the first 30 epochs. For the remaining 30 epochs, for testing purposes we conducted three separate training sessions with $w_i$ set to $100.0$, $200.0$, and $300.0$, respectively. In the end, we obtained three sets of trained modules which hereafter we refer to as $NC_{100}$, $NC_{200}$, and $NC_{300}$. Over the course of training, the encoder learns to generate Ninja Codes with increasingly subtle visual alterations (Figure 4, bottom) --- however, remnants of the multicolor stripes of earlier Ninja Codes still remain as visual artifacts. (Expectedly, training with higher $w_i$ values results in less salient artifacts.)

\begin{figure}[t]
  \centering
  \includegraphics[width=1.0\linewidth]{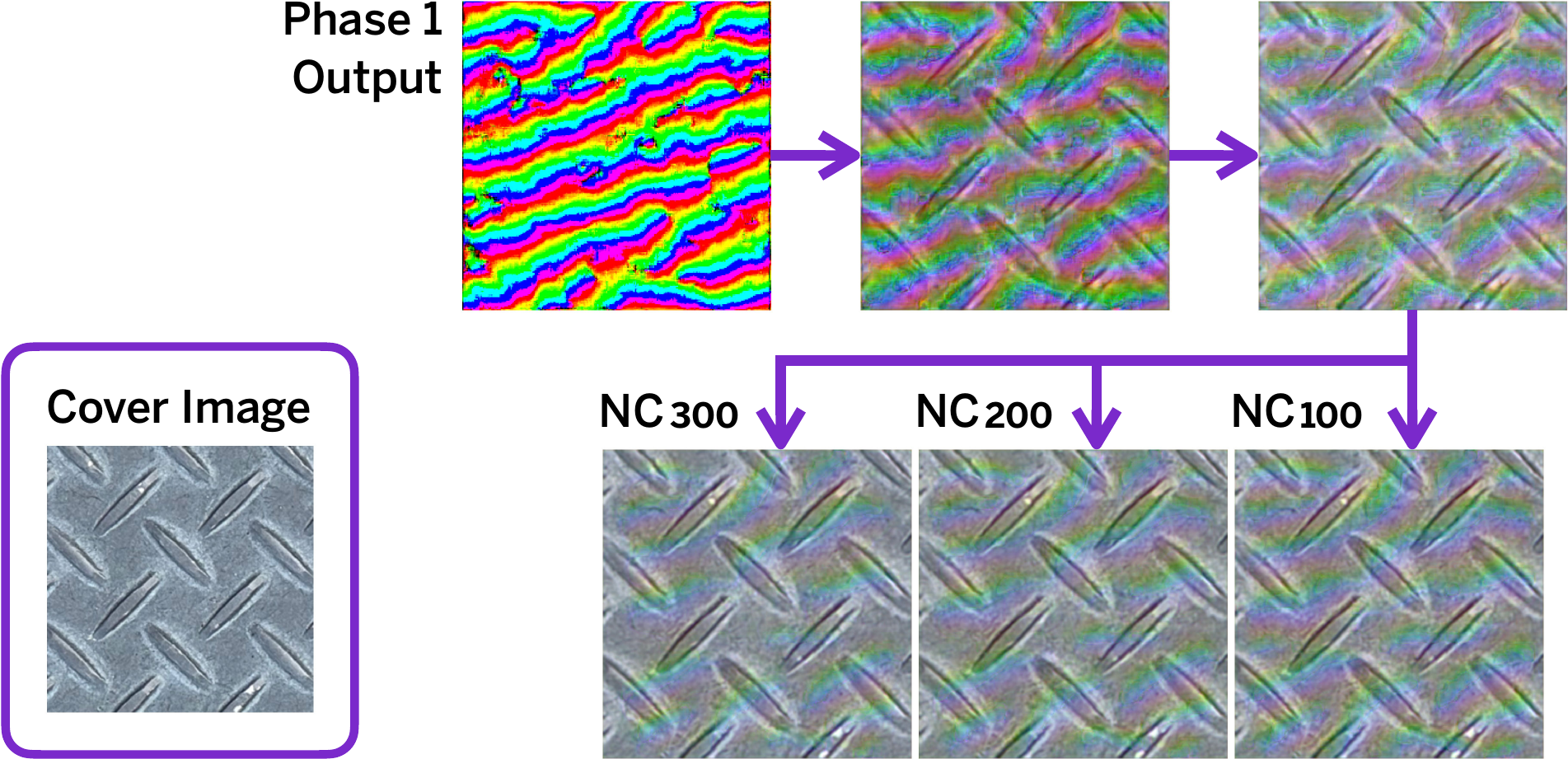}
  \caption{After the first phase of training, the encoder will produce Ninja Codes with colorful patterns (top left). The patterns fade but remain as visual artifacts after the second phase (bottom). Artifact saliency can be controlled by adjusting hyperparameters.}
\end{figure}

OPTIONAL FINE-TUNING. Due to our use of an end-to-end training process, Ninja Codes generated using a particular encoder can only be expected to be detected by modules co-trained in the same session. It would be convenient, however, if we could train multiple encoders using a variety of parameters (e.g., different $w_i$ values) to cater to different scenarios --- for example, opt for stronger artifact suppression if the codes are to be pasted onto light-colored surfaces --- and train a single set of detector/decoder modules that can handle codes generated by any of the encoders. To test whether this works in practice, we took the detector/decoder modules from $NC_{100}$ and performed 30 additional epochs of fine-tuning, to make them able to detect codes generated by any of the three encoders.

\section{Evaluation}

We conducted a series of experiments to assess the performance of Ninja Codes.

\subsection{Code Detection Performance Test}

First, to test the accuracy of region/corner detection and ID (message) retrieval, we prepared 25 high-resolution digital images of common environmental textures such as marble, wood, asphalt, and converted the central portion of each image into a Ninja Code using the $NC_{100},$ $NC_{200}$, and $NC_{300}$ encoders (Figure 5, left). We displayed each image on a 22-inch LG UltraFine 5K LCD monitor (scaled so that the Ninja Code appeared at a size of 8.5$\times$8.5cm on the screen) in a well-lit indoor room, and took a series of photographs from eight camera positions (Figure 5, right) using a custom Swift application running on an Apple iPhone 15 Pro. The photos were shot in 1080$\times$1080 resolution, and at each of the eight positions, camera orientation was adjusted so that the center of the LCD screen coincided with the center of the captured photo. The photos were given as input to the trained detector/decoder modules (running locally on the iPhone), to locate the Ninja Codes, obtain their corner coordinates, and retrieve their IDs.

\begin{figure}[t]
  \centering
  \includegraphics[width=1.0\linewidth]{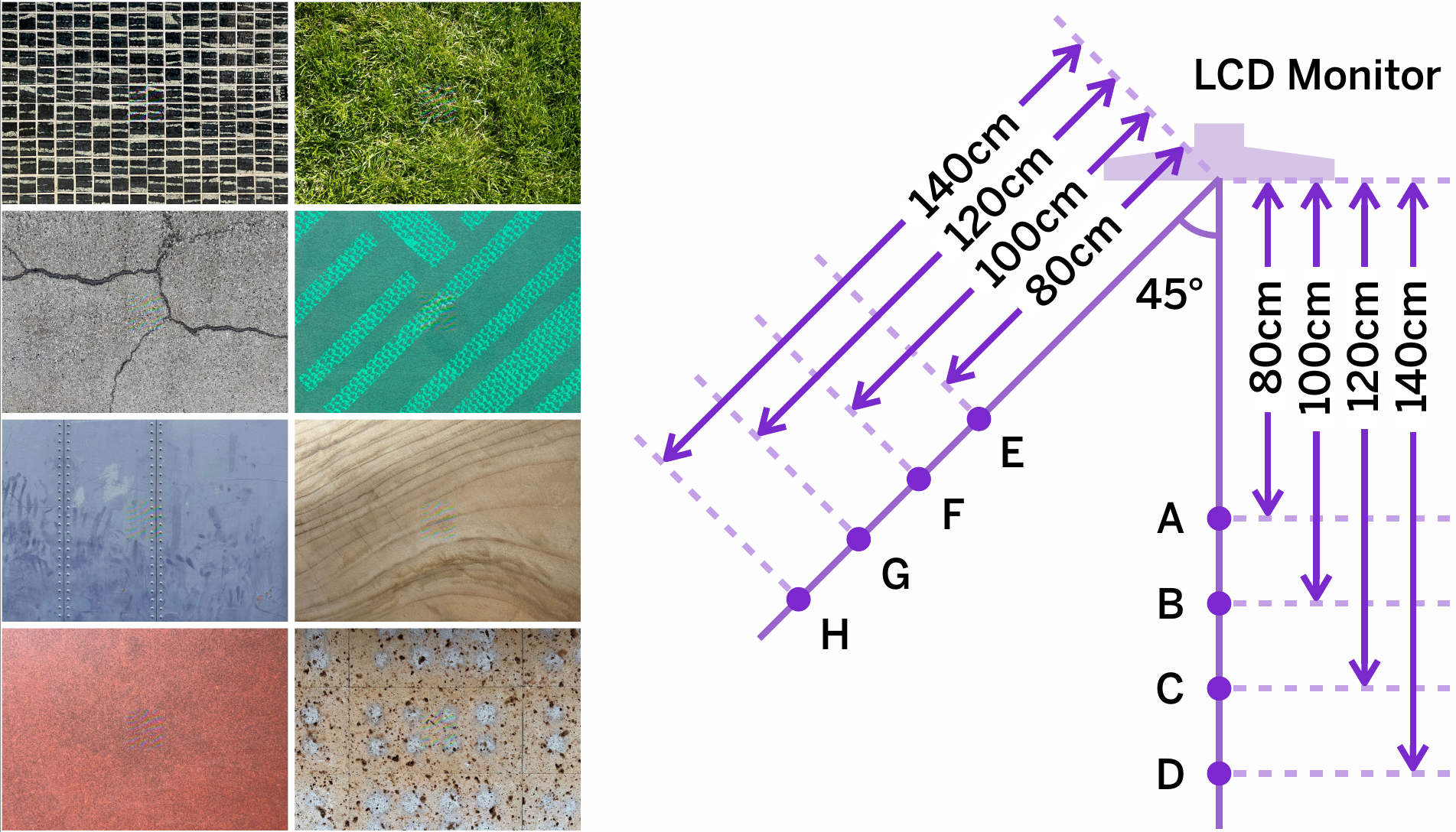}
  \caption{Examples of digital images used to evaluate code detection performance, each with a single Ninja Code (here, generated using the $NC_{300}$ encoder) placed at the center (left). The eight camera positions, as seen from above (right).}
\end{figure}

Our goal with this test was to measure the average corner error (in pixels) and drop rate, i.e., the percentage of failed detections. From each of the eight camera positions, the app took five photos at one-second intervals. The drop rate was measured by counting the number of photos for which the modules either failed to find the Ninja Code, or found the code but failed to retrieve the 36-bit ID in full. The corner error was measured using the first photo in which a Ninja Code was successfully detected with the ID intact; the predicted corner coordinates were compared with ground truth values obtained beforehand for the eight camera positions. In cases where the app failed to detect codes in all five photos, we recorded the drop rate for that position as $100\%$ and let the app continue to take photos until a successful detection occurred, from which we measured the corner error.

We conducted the test three separate times, for each of the three sets ($NC_{100}$,  $NC_{200}$, $NC_{300}$) of trained modules. In addition, we conducted the test three more times using the fine-tuned detector/decoder modules (discussed in section 3.5, hereafter abbreviated as $FT$ modules) trained to detect codes generated by any of the three encoders. Lastly, for comparison we conducted the test with ArUco markers used in place of Ninja Codes.

Table 1 shows the results of this test. The average corner error hovers around one pixel, which is larger than that of ArUco but still small in absolute terms. Notably, we did not see significant differences in errors between the eight camera positions. Also, the fine-tuned $FT$ modules yielded results similar to those for dedicated modules, suggesting the feasibility of preparing a suite of Ninja Codes (each catering to different usage scenarios) which can be detected by a single set of detector/decoder modules. (Note: While all photos were shot in 1080$\times$1080 resolution, the region detector accepts 300$\times$300 images and thus performs a scaling operation before processing. Values in Table 1 are reported in that scale and should be interpreted accordingly. To ensure fair comparison, ArUco markers were also detected from similarly scaled down images.)

The results also show high drop rates for Ninja Codes, especially for those created using the $NC_{300}$ encoder. We noticed that failed detections were background-specific --- they did not occur at similar rates for all texture images, but were instead concentrated around those with high visual contrast. For each image, we quantified its contrast intensity using local RMS (standard deviation of pixel YUV values, computed at 32$\times$32 patches throughout the image and then averaged); the two images with the highest RMS (the tile and grass images shown on the top row in Figure 5, left) coincided with those that produced the most failed detections. Once we omit these two images from our analysis, the drop rate decreased to 1.30\% for $NC_{100}$, 4.67\% for $NC_{200}$, and 8.15\% for $NC_{300}$. This fact may possibly be used to caution users when they attempt to generate Ninja Codes from cover images with overly high visual contrast.

\begin{table}[t]
  \caption{Code Detection Performance Results}
  \begin{tabular}{ccc}
    \toprule
    &Corner Error (px)&Drop Rate (\%)\\
    \midrule
    $NC_{100}$&0.994&3.20\\
    $NC_{200}$&1.057&7.30\\
    $NC_{300}$&1.145&11.10\\
    $NC_{100}$ + $FT$&0.958&3.50\\
    $NC_{200}$ + $FT$&1.042&5.90\\
    $NC_{300}$ + $FT$&1.220&12.80\\
    ArUco&0.586&0.00\\
    \bottomrule
  \end{tabular}
\end{table}

Furthermore, we found that the majority of failed detections were caused by faulty message retrieval, i.e., the Ninja Code had been successfully located but the decoder failed to recover the 36-bit ID in full. This suggests that the drop rate may be reduced by employing error correction, using part of the 36-bit string as check symbols. To test this, we prepared an alternative set of $NC_{300}$ codes where 12 of the 36 bits were designated as check symbols, and implemented Reed-Solomon error correction to repair corrupt bits (if any) after decoding. While this has the effect of decreasing the number of unique IDs that can be represented (to approximately 16.8 million from 68.7 billion), subsequent testing revealed a sizable reduction in drop rate, to 6.00 from 11.10.

Overall, the results show high detection performance for Ninja Codes, with several caveats: the tendency to struggle with high-contrast texture images, and the possible need for error correction if opting for strong artifact suppression.

\begin{figure}[t]
  \centering
  \includegraphics[width=1.0\linewidth]{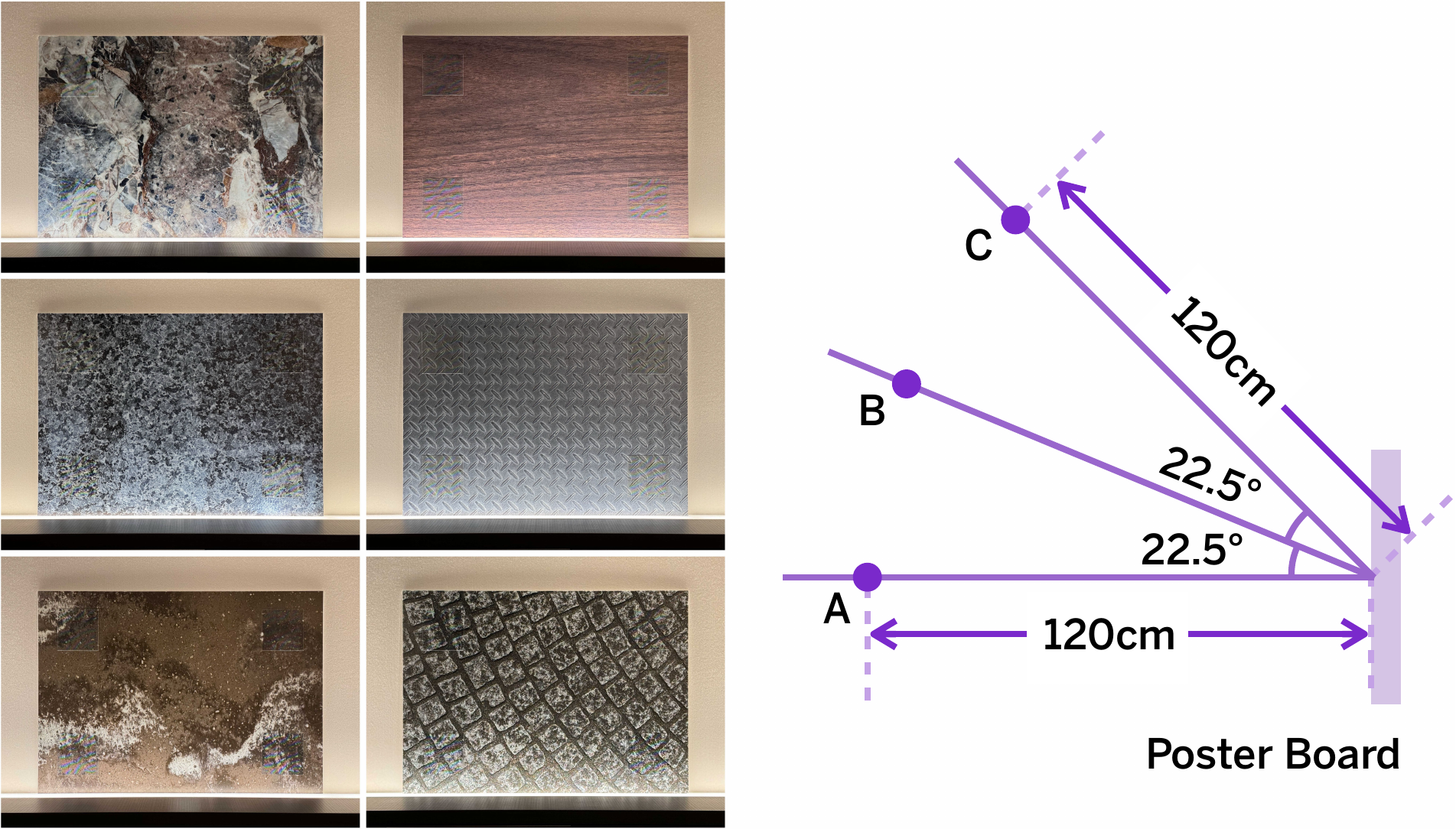}
  \caption{Examples of poster boards used to evaluate 6-DoF tracking performance, with four Ninja Codes (generated using the $NC_{200}$ encoder) placed at the corners (top). The three camera positions, as seen from above (bottom).}
\end{figure}

\subsection{6-DoF Tracking Performance Test}

Next, we prepared ten A2 size (59.4$\times$42.0cm) poster boards (Figure 6, left), each showing a texture image selected from the 25 images used in the previous test.  (Out of the 25 images, we excluded the two with the highest local RMS and randomly selected ten from the rest.) For each board, we attached four 8.5$\times$8.5cm Ninja Codes (generated using the mid-level $NC_{200}$ encoder) to its corners, and shot a series of ten photographs from each of three camera positions (Figure 6, right). For comparison, we also conducted the same test using ArUco markers instead of Ninja Codes.  Again, all photos were shot using our custom app on an iPhone 15 Pro, in 1080$\times$1080 resolution. Both the poster boards and Ninja Codes were printed using an HP DesignJet Z9 printer. The test was conducted in a well-lit indoor environment.

For this test, in addition to detecting individual codes and retrieving their IDs, we used prior knowledge of code position/dimensions to calculate the 6-DoF position and pose of the iPhone camera. (All processing was done locally on the iPhone 15 Pro.) Here, our goal was to assess how much the corner detection accuracy of Ninja Codes (which we found to be slightly lower compared to that of ArUco) affects their 6-DoF tracking performance. In addition, we aimed to see if the difference in materials (codes were printed on paper this time) leads to any negative effects, and if there are any issues with tracking multiple Ninja Codes simultaneously.

Table 2 shows the results. We can see that Ninja Codes, while not quite matching the performance of ArUco, yield centimeter-level (position) and nearly single-degree (pose) accuracy which should be sufficient for a wide range of use cases. We did not observe any issues regarding multi-code detection, nor did we observe adverse effects attributable to printing method/material.

\subsection{Visual Conspicuity Test}

To evaluate how well Ninja Codes can conceal themselves from human detection, we conducted a software-based test using a custom application built for the Microsoft Surface Pro 7 touchscreen laptop. The app shows a full-screen rendering of a texture image (randomly sampled from DTD), onto which a single Ninja Code is overlaid with randomized size and location. Tapping on the Ninja Code triggers the software to record the time passed since the image was displayed; tapping anywhere else on the screen does nothing. We recruited nine test subjects in a university setting (all male, average age: 23.4), and measured the time required for them to locate Ninja Codes on the app. For comparison, we built a separate version of the app where ArUco markers were overlaid instead of Ninja Codes. Each subject was shown 25 code-overlaid images for both versions of the app, and for each image was asked to find and tap on the code as quickly as possible. Due to scheduling and logistical issues, Ninja Codes in this test were created using a prototype version of the encoder which output codes with stronger visual artifacts compared to the newer $NC_x$ encoders (see supplementary material for images).

\begin{table}[t]
  \caption{6-DoF Tracking Performance Results}
  \begin{tabular}{ccc}
    \toprule
    &Position Error (cm)&Pose Error (degrees)\\
    \midrule
    $NC_{200}$&2.988&1.084\\
    ArUco&2.121&0.854\\
     \bottomrule
  \end{tabular}
\end{table}

\begin{figure}[b]
  \centering
  \includegraphics[width=1.0\linewidth]{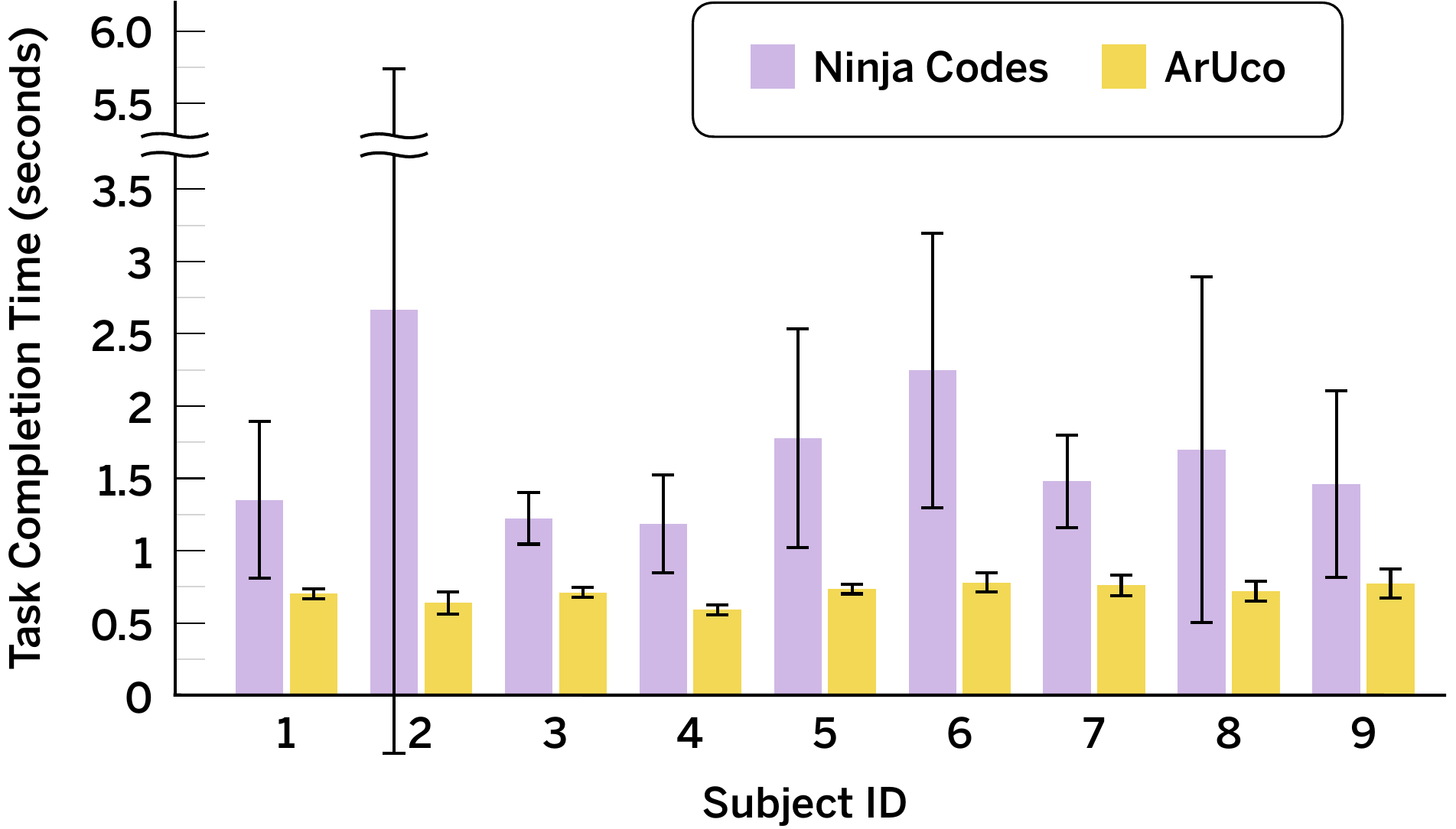}
  \caption{Average code detection time for Ninja Codes and ArUco markers. Error bars represent 95\% confidence intervals.}
\end{figure}

\begin{figure}[t]
  \centering
  \includegraphics[width=1.0\linewidth]{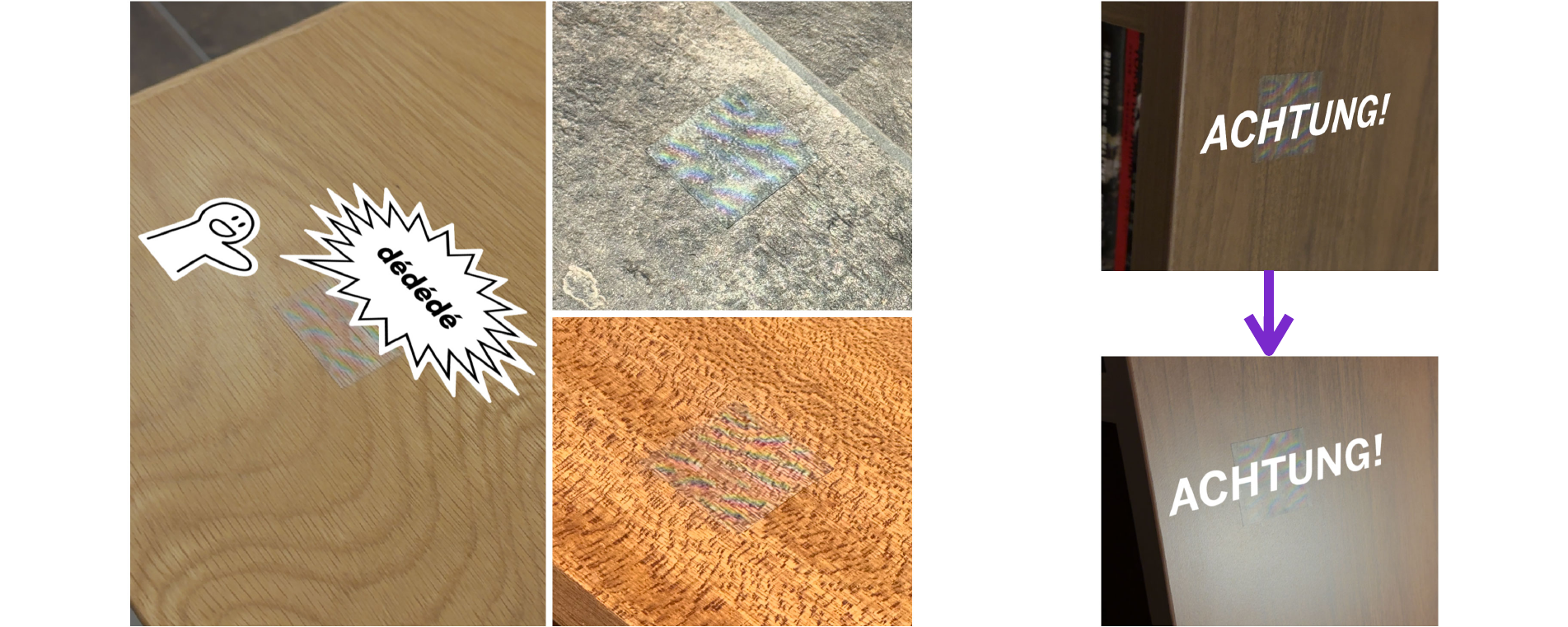}
  \caption{A rudimentary augmented reality app and printed Ninja Codes (left) used for informal observations. The codes are reasonably robust to changes in lighting conditions (right).}
\end{figure}

Figure 7 shows the results. Subjects needed significantly more time to locate Ninja Codes, compared to ArUco markers which were generally detected promptly. Detection time was also highly variable for Ninja Codes, with high-contrast texture images causing particular trouble for subjects --- interestingly, the same types of images that yielded high drop rates in the first test.

\subsection{Informal Observations}

In addition to the above quantitative evaluations, to gain insights regarding how Ninja Codes may perform in real-life scenarios, we extended our iPhone app giving it a rudimentary augmented reality function and informally used it in an indoor environment where a series of Ninja Codes had been installed beforehand (Figure 8, left). The codes were generated using the $NC_{300}$ encoder, and printed using a Fujifilm Multifunction printer as 4.25$\times$4.25cm squares.

As a result, our experiences largely conformed to prior expectations. One issue we found was that material differences between the Ninja Codes and background surfaces often made the codes easily noticeable, especially for surfaces with three-dimensional textures such as textiles. Regarding detection performance, we found robust performance under a wide variety of lighting conditions, including cases where the codes were partially obscured due to specular reflection (Figure 8, right) --- suggesting that data is encoded within the codes in a spatially redundant manner, as per our intention in designing the noise functions. We also noticed that in general, detection performance deteriorates less for underilluminated situations than overilluminated ones. Although we suspect that this has more to do with the iPhone's auto-exposure algorithm than Ninja Codes per se, a small adjustment in camera noise functions may work as a remedy.

\section{Limitations and Extensions}

Below, we list several outstanding issues (along with countermeasures) and avenues for further development regarding our work. (See supplementary material for more details.)

VISUAL ARTIFACTS. Ninja Codes, despite concealing themselves well from human detection, still produce visual artifacts which manifest particularly strongly for codes created from plain or light-colored cover images. Further suppression should be possible via model refinements, preparation of dedicated texture datasets, etc. An alternative strategy may be to seek aesthetic control of artifacts by updating loss functions, i.e., to accept the presence of artifacts but ensure they will stay visually unobjectionable.

MAXIMUM RANGE. Our region detector is designed to take 300$\times$300 images as input; at this resolution, codes with smaller dimensions, or those located at large distances, fail to retain enough visual features to be detected reliably. (Under the experimental setup  in section 4.1, the maximum distance from which our modules could robustly detect the codes was approximately 160cm.) A straightforward solution to this would be to modify the detector to accept larger images, which may come with the byproduct of potentially slower training and inference. For reference, our current detector/decoder modules have a combined inference time of around 30 ms on the iPhone 15 Pro.

PRIVACY. Location-tracking technology inevitably entails privacy concerns. To give an example, if a person takes a photo in an environment with Ninja Codes and uploads it to the internet, an attacker may be able to extract precise location information even if metadata had been stripped from the file. A possible defense --- albeit an incomplete one --- may be mounted against such attacks, by exploiting the fact that Ninja Codes modules are interoperable only with other modules co-trained in the same session; hence we can create ``private'' versions of Ninja Codes, and only permit trusted parties to access the corresponding detectors/decoders.

REVERSE ENCODER. An interesting finding we made through our experiments is that a standard U-Net model can be trained to act as a {\it reverse encoder}, which takes a Ninja Code as input and attempts to reconstruct the cover image. This may be useful, for example, in augmented reality applications where we wish to remove all hints of Ninja Codes from the final augmented scene, such as when using them in a film production setting as anchors to position CGI elements in space. The reverse encoder may also serve as a defense against the aforementioned privacy issue, by enabling Ninja Codes to be automatically obscured in photographs to prevent surreptitious extraction of location data.

COLOR CALIBRATION. Accurate color calibration is critical to curb visual discontinuities between Ninja Codes and their background surfaces. While specialized hardware and techniques exist \cite{hawkins2001} for such purposes, deploying such solutions in real-life scenarios is often impractical. For our experiments, we have created a simple tool (see supplementary material for images) that assists calibration by capturing photos under controlled LED lighting. Further development of such usability-enhancing tools will be imperative for wide adoption of the technology.

MATERIALITY. So far, we have developed Ninja Codes with the assumption that the codes will primarily be printed on paper. However, the underlying concept is theoretically adaptable to various physical media whose fabrication processes can be digitally controlled, such as 3D printed objects and textiles. Expanding materiality to better reflect the physical heterogeneity of the world will give us a wider set of options for concealing Ninja Codes in environments; we view this as an important direction of future work.

\section{Conclusion}

In this paper we described Ninja Codes, neurally generated fiducial markers that offer reliable 6-DoF location tracking while blending seamlessly into various real-world environments. With further refinements, we foresee the technology as providing the necessary foundation for numerous applications that rely on precision location tracking to make their ways into new contexts.
{
    \small
    \bibliographystyle{ieeenat_fullname}
    \bibliography{references}
}

\clearpage
\setcounter{page}{1}
\maketitlesupplementary

\section{Project Website}

Video  and other additional material regarding our work can be found on our project website, at the following URL:

\noindent
\href{https://sento.net/research/ninjacodes}{\normalsize\tt https://sento.net/research/ninjacodes}

\section{Ninja Code Examples}

Figures 9 and 10 are high-resolution images of Ninja Codes used in our experiments, in digital and printed forms.

\begin{figure}[h]
  \centering
  \includegraphics[width=1.0\linewidth]{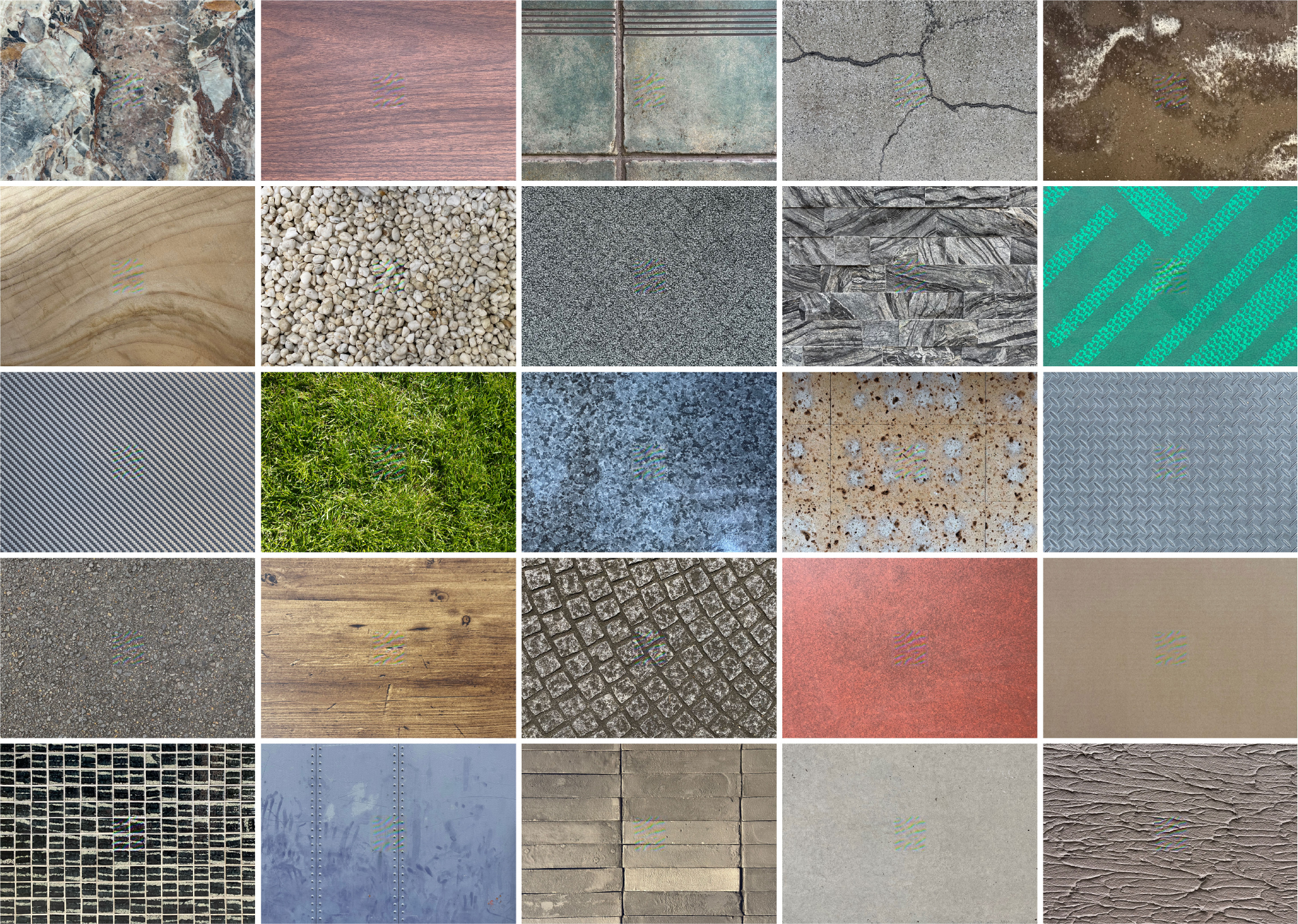}
  \caption{The 25 digital images used to evaluate code detection performance, each with a single Ninja Code (here, generated using the $NC_{300}$ encoder) placed at the center.}
\end{figure}

\begin{figure}[h]
  \centering
  \includegraphics[width=1.0\linewidth]{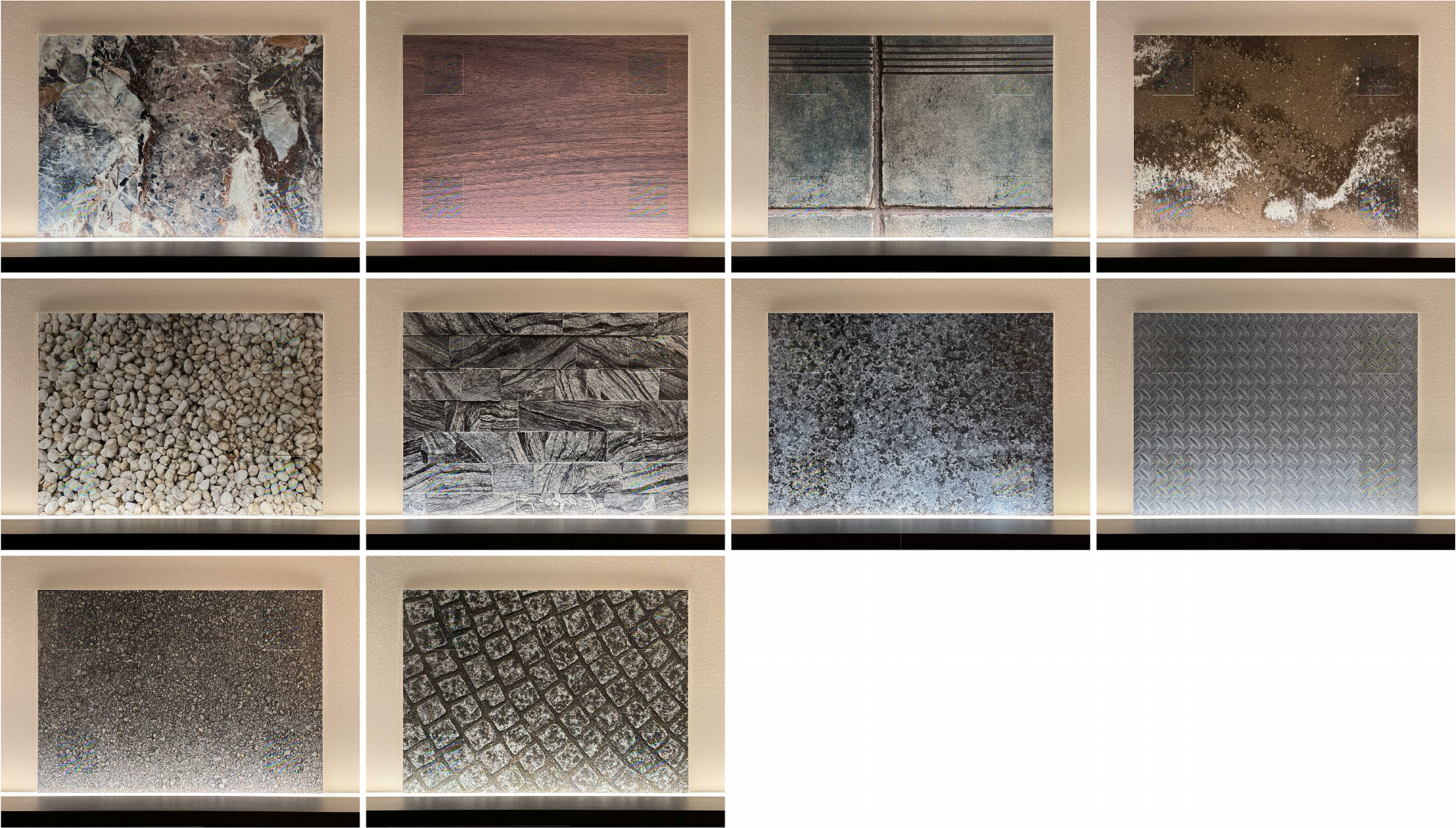}
  \caption{The ten poster boards used to evaluate 6-DoF tracking performance, with four Ninja Codes (generated using the $NC_{200}$ encoder) placed at the corners.}
\end{figure}

\section{Prototype Codes}

Figure 11 shows several examples of prototype Ninja Codes used in the visual conspicuity test.

\begin{figure}[h]
  \centering
  \includegraphics[width=1.0\linewidth]{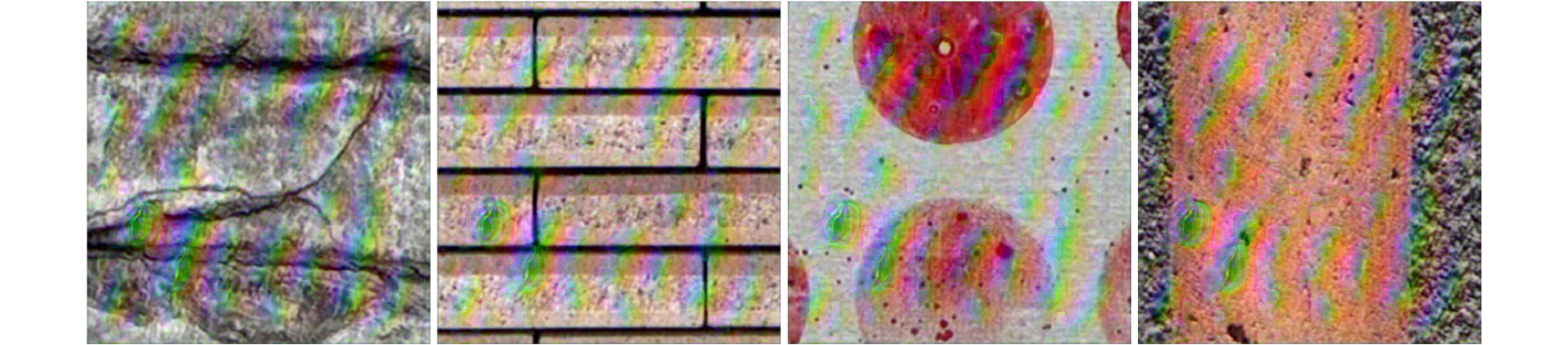}
  \caption{Prototype Ninja Codes used to assess visual conspicuity, exhibiting more salient visual artifacts compared to the newer $NC_x$ encoders.}
\end{figure}

\section{Extensions Details}

Figure 12 shows the reverse encoder reconstructing original cover images from Ninja Codes, albeit with some loss of fine detail. Figure 13 shows our custom photographic tool used to assist color calibration. (The codes in Figure 8 were created using this tool.)

\begin{figure}[h]
  \centering
  \includegraphics[width=1.0\linewidth]{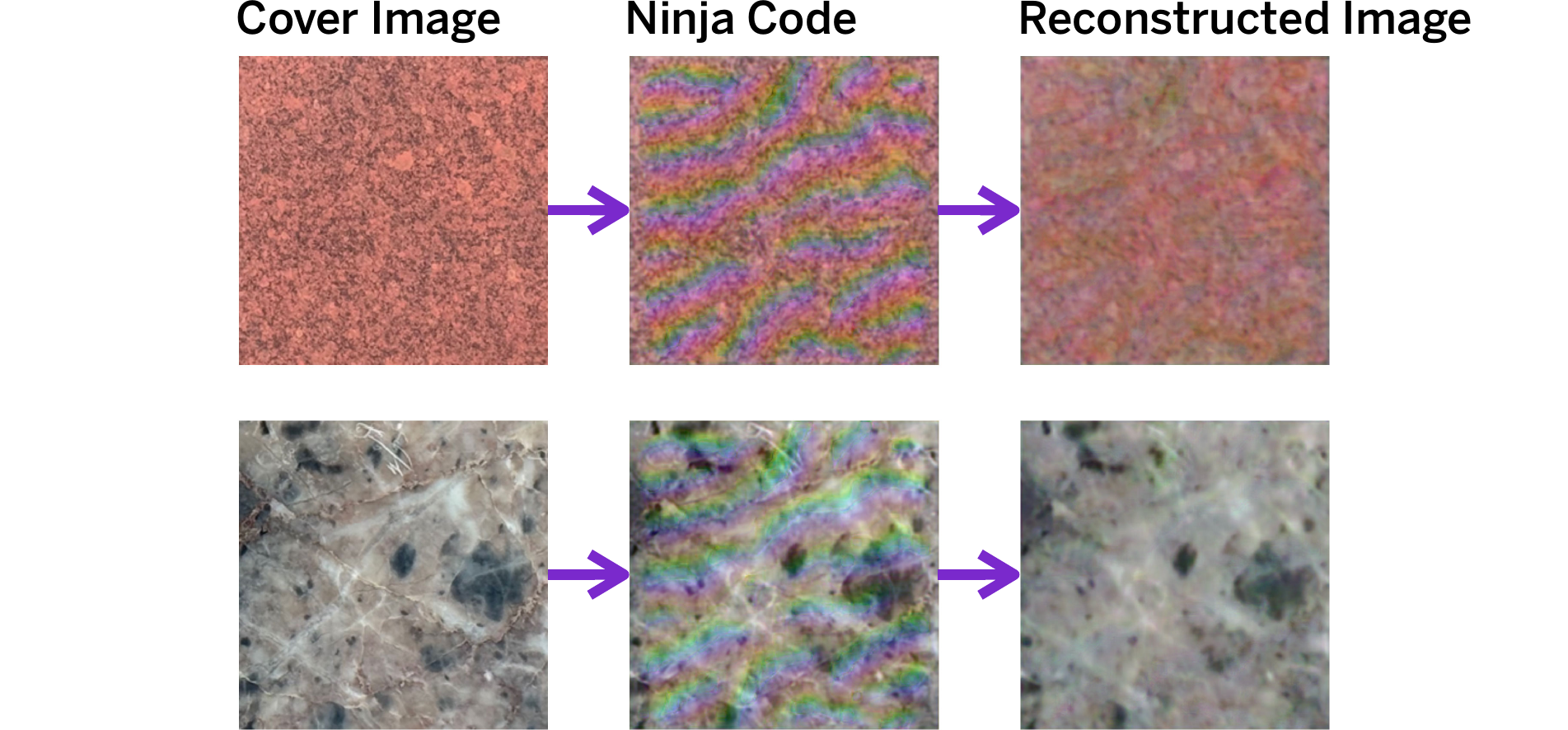}
  \caption{A U-Net based reverse encoder that takes a Ninja Code as input and attempts to reconstruct the cover image.}
\end{figure}

\begin{figure}[h]
  \centering
  \includegraphics[width=1.0\linewidth]{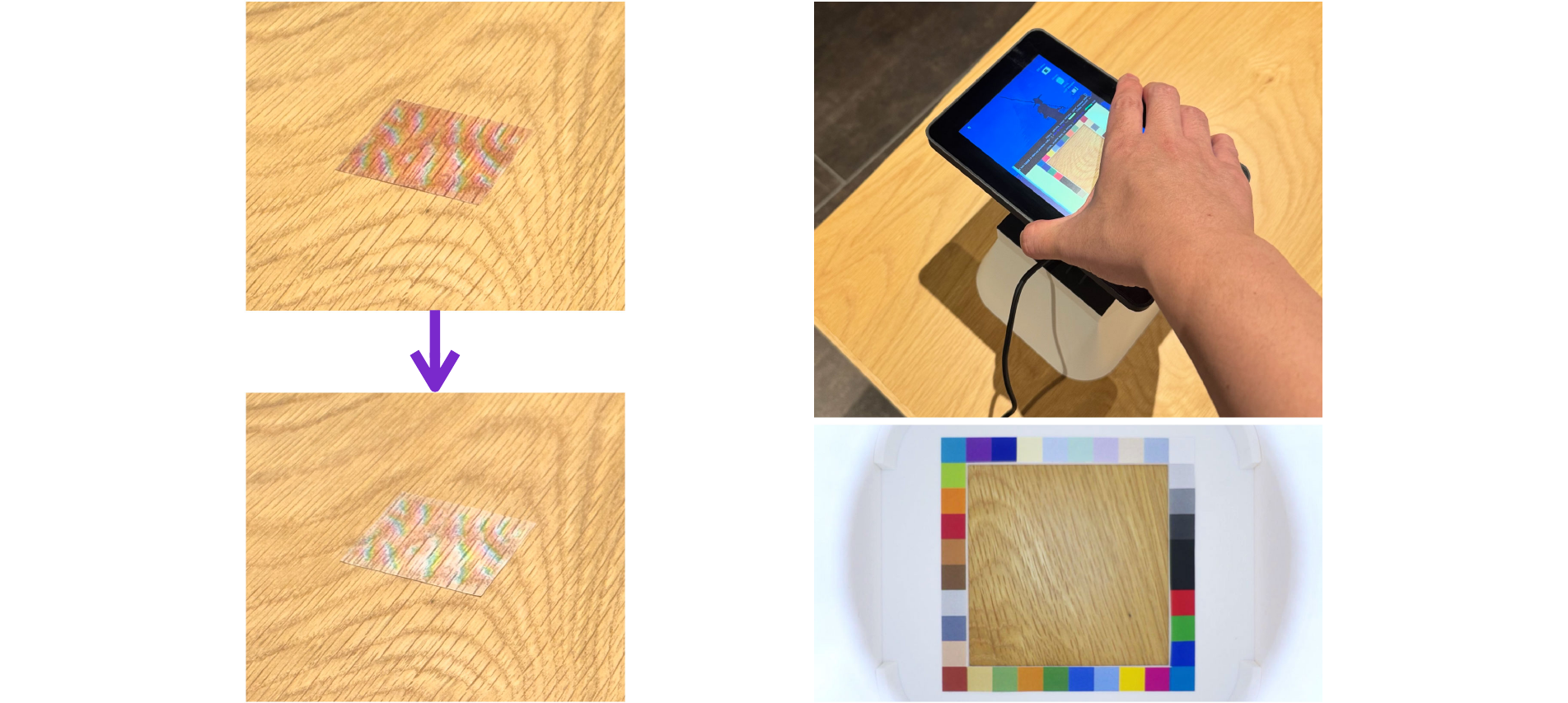}
  \caption{Faulty color calibration resulting in color discontinuity (left). A photographic tool that captures texture images under controlled LED lighting (right).}
\end{figure}

\end{document}